# Probability Update: Conditioning vs. Cross-Entropy

**Adam J. Grove**
NEC Research Institute
4 Independence Way
Princeton, NJ 08540
grove@research.nj.nec.com

**Joseph Y. Halpern**
Cornell University
Dept. of Computer Science
Ithaca, NY 14853
halpern@cs.cornell.edu
http://www.cs.cornell.edu/home/halpern

## Abstract

Conditioning is the generally agreed-upon method for updating probability distributions when one learns that an event is certainly true. But it has been argued that we need other rules, in particular the rule of cross-entropy minimization, to handle updates that involve uncertain information. In this paper we re-examine such a case: van Fraassen's *Judy Benjamin* problem [1987], which in essence asks how one might update given the value of a conditional probability. We argue that—contrary to the suggestions in the literature—it is possible to use simple conditionalization in this case, and thereby obtain answers that agree fully with intuition. This contrasts with proposals such as cross-entropy, which are easier to apply but can give unsatisfactory answers. Based on the lessons from this example, we speculate on some general philosophical issues concerning probability update.

## 1  INTRODUCTION

How should one update one's beliefs, represented as a probability distribution Pr over some space $S$, when new evidence is received? The standard Bayesian answer is applicable whenever the new evidence asserts that some event $T \subseteq S$ is true (and furthermore, this is *all* that the evidence tells us). In this case we simply condition on $T$, leading to the distribution $\Pr(\cdot|T)$.

For successful "real-world" applications of probability theory so far, conditioning has been a mostly sufficient answer to the problem of update. But many people have argued that conditioning is not a *philosophically* adequate answer (in particular, [Jeffrey 1983]). Once we try to build a truly intelligent agent interacting in complex ways with a rich world, conditioning may end up being practically inadequate as well.

The problem is that some of the information that we receive is not of the form "$T$ is (definitely) true" for any $T$. What would one do with a constraint such as "$\Pr(T) = 2/3$" or "the expected value of some random variable on $S$ is $2/3$". We cannot condition on this information, since it is not an event in $S$. Yet it is clearly useful information. So how should we incorporate it? There is in fact a rich literature on the subject (e.g., see [Bacchus, Grove, Halpern, and Koller 1994; Diaconis and Zabell 1982; Jeffrey 1983; Jaynes 1983; Paris and Vencovska 1992; Uffink 1995]). Most proposals attempt to find the probability distribution that satisfies the new information and is in some sense the "closest" to the original distribution Pr. Certainly the best known and most studied of these proposals is to use the rule of minimizing *cross-entropy* [Kullback and Leibler 1951] as a way of updating with general probabilistic information. This rule can also be shown to generalize *Jeffrey's rule* [Jeffrey 1983], which in turn generalizes conditioning.

But is cross-entropy (CE) really such a good rule? The traditional justifications of CE are that it satisfies various sets of criteria (such as those of [Shore and Johnson 1980]) which, while plausible, are certainly not compelling [Uffink 1995]. Van Fraassen, in a paper entitled "A problem for relative information [CE] minimizers in probability kinematics" [1981] instead approached the question in a different way: he looked at how CE behaves on a simple specific example. He calls his example the *Judy Benjamin* (JB) problem; in essence it is just the question of how one should update by a *conditional* probability assertion, i.e., "$\Pr(A|B) = c$" for some events $A, B$ and $c \in [0, 1]$.

As we now explain, van Fraassen uncovers what seems (to us) to be an unintuitive feature of cross-entropy, although in later papers on the same issue he endorses CE and a family of other similar rules. Furthermore, none of his rules agree with most people's strong intuition about the solution to his problem. The purpose of this paper is to give a new analysis, which is based on simple conditionalization, and is (we argue) in good agreement with people's expectations. Our hope is that the example and our analysis will be an instructive study of the subtleties involved in probability update, and in particular the dangers involved in indiscriminately applying supposedly "simple" and "general" rules like CE.

Van Fraassen explains the JB problem as follows [1987]:

> [The story] derives from the movie *Private Benjamin*, in which Goldie Hawn, playing the title character, joins the Army. She and her platoon, participating in war games on the side of the "Blue Army", are dropped in the wilderness, to scout the opposition ("Red Army"). They are soon lost. Leaving the movie script now, suppose the area is divided into two halves, Blue and Red territory, which each territory is divided into Headquarters Company area and Second Company area. They were dropped more or less at the center, and therefore feel it is equally likely that they are now located in one area as in another. This gives us the following muddy Venn diagram, drawn as a map of the area:

| 1/4 Red 2nd | 1/4 Red HQ |
|:---|:---|
| 1/2 Blue | |

> They have some difficulty contacting their own HQ by radio, but finally succeed and describe what they can see around them. After a while, the office at HQ radios: "I can't be sure where you are. If you are in Red territory, the odds are 3:1 that you are in HQ Company area ..." At this point the radio gives out.

We must now consider how Judy Benjamin should adjust her opinions, if she accepts this radio message as the sole and correct constraint to impose. The question on which we should focus is: what does it do to the probability that they are in friendly Blue territory? Does it increase, or decrease, or stay at its present level of 1/2?

The intuitive response is that the message should not change the *a priori* probability of 1/2 of being in Blue territory. More precisely, according to this response, Judy's posterior probability distribution should place probability 1/4 on being at each of the two quadrants in the Blue territory, probability 3/8 on being in the Headquarters Company area of Red territory, and probability 1/8 on being in the Second Company area of Red territory. Van Fraassen [1987] notes that his many informal surveys of seminars and conference audiences find that people overwhelmingly give this answer.

However, this intuitive answer is inconsistent with cross-entropy. In fact, it can be shown that cross-entropy has the rather peculiar property that if HQ had said "If you are in Red territory, the odds are $\alpha : 1$ that you are in HQ company area ...", then for all $\alpha \neq 1$, the posterior probability of being in Blue territory (according to the distribution that minimizes cross-entropy and satisfies this constraint) would be *greater* than 1/2; it would stay at 1/2 only if $\alpha = 1$. This seems (to us, at least) highly counterintuitive. Why should Judy come to believe she is more likely to be in Blue territory, almost no matter what the message says about $\alpha$? For example, what if Judy knew in advance that she would receive such a message for some $\alpha \neq 1$, and simply did not know the value of $\alpha$. Should she then increase the probability of being in Blue territory even before hearing the message? As van Fraassen [1981] says, as part of an extended discussion of this phenomenon:

> It is hard not to speculate that the dangerous implications of being in the enemy's headquarters area are causing Judy Benjamin to indulge in wishful thinking...[1]

However, as van Fraassen points out (crediting Peter Williams for the observation), there also seems to be a problem with the intuitive response. Presumably there is nothing special about hearing the odds of being in Red territory are 3:1. The posterior probability of being in Blue territory should be 1/2 no matter what the odds are, if we really believe that this information is irrelevant to the probability of being in Blue territory. But if $\alpha = 0$ then, to quote van Fraassen [1987], "he would have told her, in effect, 'You are *not* in Red Second Company area'". Assuming that this is indeed equivalent, it seems that Judy could have used simple conditionalization, with the result that her posterior probability of being in Blue territory would be 2/3, not 1/2.

In [van Fraassen 1987; van Fraassen, Hughes, and Harman 1986], van Fraassen and his colleagues formulate various principles that they argue an update rule should satisfy. Their first principle is motivated by the observation above and simply says that, when conditioning seems applicable, the answer should be that obtained by conditioning. To state this more precisely, let $q = \alpha/(1+\alpha)$ be the probability (rather than the odds) of being in red HQ company area given that Judy is in Red territory. In the case of the JB problem, the first principle becomes:

> If $q = 1$ the prior is transformed by simple conditionalization on the event *"Red HQ area or Blue territory"*; if $q = 0$ by simple conditionalization on *"Red 2nd company area or Blue territory"*.

---

[1] We remark that this behavior of CE has also been discussed and criticized in a more general setting [Seidenfeld 1987].

This first principle already eliminates the intuitive rule, i.e., the rule that the posterior probability should stay at $1/2$ no matter what $q$ is. (Note also that we cannot make the rule consistent with this principle by treating $q = 0$ and $q = 1$ as special cases, unless we are prepared to accept a rule that is discontinuous in $q$.) For van Fraassen, this is apparently a decisive refutation of the intuitive rule, which he thus says is flawed [van Fraassen 1987].

However, in this paper we give a new[2] and simple analysis of the JB problem. We believe that our solution is well-motivated, and it agrees completely with the intuitive answer. It thus also does not exhibit the counterintuitive behavior of CE.

Our basic idea is simply to use conditioning, but to do so in a larger space where it makes sense (i.e., where the information we receive *is* an event). Of course, people have always realized that this option is available. It is perhaps not popular because it appears to pose certain serious philosophical and practical problems as a general approach. In particular, *which* larger space do we use? There may be many equally natural possibilities, leading to different answers, so the rule will be indeterminate. Also possible is that all extended spaces we can think of seem equally unnatural and contrived; again, we will be stuck. In addition, there is the practical concern that a rich enough space might be vastly larger and more complicated to work in than the original.

Against this, a rule like cross-entropy seems extremely attractive. It provides a single general recipe which can be mechanically applied to a huge space of updates. Even families of rules, such as van Fraassen proposes, are not so bad: after one has chosen a rule (usually by selecting a single real-valued parameter [Uffink 1995]), the rest is again mechanical, general, and determinate. Furthermore, all these rules work in the original space $S$, without requiring expansion, and so may be more practical in a computational sense.

Since all we do in this paper is analyze one particular problem, we must be careful in making general statements on the basis of our results. Nevertheless, they do seem to support the claim that sometimes, the only "right" way to update, especially in an incompletely specified situation, is to think very carefully about the real nature and origin of the information we receive, and then (try to) do whatever is necessary to find a suitable larger space in which we can condition. If this doesn't lead to conclusive results, perhaps this is because we do not understand the information well enough to update with it. However much we might wish for one, a *generally* satisfactory mechanical rule such as cross-entropy, which saves us all this questioning and work, probably does not exist.

This does not deny the usefulness of rules like CE. There will be some (perhaps large) family of situations in which CE is indeed appropriate, and we would like to better understand what this family is and how to recognize it. But if CE (or any other rule) is blindly applied *whenever* the information is of the appropriate syntactic form, we should not be surprised if the results are often unexpected and unhelpful.

## 2   CONDITIONING

In this section, we present our alternative analysis of the JB problem. In the following, let $B_1$, $B_2$, $R_1$, $R_2$ denote the events that Judy is in, respectively, Blue HQ, Blue Second Company, Red HQ, and Red Second Company areas. Let $B = B_1 \vee B_2$ and $R = R_1 \vee R_2$. The message HQ sent, "If you are in Red territory, the odds are 3:1 that you are in HQ Company area", is equivalent to asserting that the conditional probability of $R_1$ given $R$ is true is 3/4. In general, let $\mathbf{M}(q)$ be the similar message asserting that this probability is $q \in [0,1]$ for some $q$ not necessarily $= 3/4$ (i.e., the announced odds are $q/(1 - q) : 1$ instead of $3 : 1$). We use $\Pr_J^{prior}$ to denote Judy's prior beliefs (i.e., before the message is received) and $\Pr_J^q$ to denote her posterior distribution after receiving $\mathbf{M}(q)$.

The key step is to re-examine the problem from the beginning, and ask ourselves how Judy should treat HQ's message. Note that Van Fraassen explicitly assumes, in his statement of the problem, that HQ's statement should be treated as a constraint on Judy's beliefs. Thus, he interprets it as an imperative: "Make your beliefs be such that this is true!" This interpretation of probabilistic information as constraints is a common one (especially in the context of CE), but is difficult to justify [Uffink 1996]. Van Fraassen is, of course, quite aware of the philosophical issues raised by his interpretation; see [van Fraassen 1980].

But is the interpretation that HQ's statement should be regarded as a constraint on Judy's beliefs the only possible one? Note that, as the story is presented, it certainly sounds as though HQ was trying to give Judy some true and useful information. But, at the time $\mathbf{M}$ is sent, the statement that $\Pr_J^{prior}(R_1|R) = 3/4$ is clearly not true of Judy's beliefs. Thus, if we wish to interpret $\mathbf{M}$ as referring directly to *Judy's* beliefs, we will be unable to regard it as a factual assertion in any straightforward sense.

But suppose we instead view $\mathbf{M}(q)$ as being a correct statement regarding *HQ's* beliefs; i.e., as asserting $\Pr_{HQ}(R_1|R) = q$ where $\Pr_{HQ}$ denotes HQ's distribution over where Judy might be. This certainly seems to be a reasonable interpretation in light of the story. How should Judy react to it? Unfortunately, the story does not give us enough information to be able to provide a definite answer to this question. Judy's correct reaction to the message depends on aspects of the situation that were not included in the problem statement.

[2]Although we are aware of no published analysis similar to our own, we have learned that Seidenfeld has earlier presented a closely related analysis in several lectures [Seidenfeld 1997].

The following is a partial list of things that could be relevant: What does Judy know about HQ's beliefs and knowledge? How did HQ expect Judy to react to the message, and what did Judy know about these expectations? What messages *could* HQ have sent? For instance, might HQ have sent $\mathbf{M}(q)$ for some other value of $q$ if it were appropriate, or would it have said something else entirely if $q \neq 3/4$? Does Judy believe that HQ even has the option of sending messages that are not of the form $\mathbf{M}(q)$; if so, what messages?[3] And so on.

We now give one particular analysis, which follows by filling in such missing details in what we feel is a plausible way. As we said, we assume that $\mathbf{M}(3/4)$ is a factual statement about HQ's beliefs. We further assume that, no matter what HQ actually knows, its message would always have been simply $\mathbf{M}(q)$ for the appropriate value of $q$. Thus, Judy can read nothing more into $\mathbf{M}(q)$ other than to regard it as a true statement about HQ's beliefs. As noted in the previous paragraph, even to get this far relies on several strong assumptions.

Once Judy hears $\mathbf{M}(q)$, it seems natural to want to condition on it. The problem is that, as yet, we have not introduced a space in which $\mathbf{M}(q)$ is an event. Of course, there are many possible such spaces. To construct an appropriate one, we must consider how Judy models HQ's beliefs, and what she believes about these beliefs. Again, here we make perhaps the simplest possible choice. We suppose that Judy models HQ's beliefs as a distribution over the four quadrants $R_1, R_2, B_1, B_2$. Let $\mathrm{Pr}_{HQ}^{a,b,c}$ be the distribution on $\{R_1, R_2, B_1, B_2\}$ such that $\mathrm{Pr}_{HQ}^{a,b,c}(R_1) = a$, $\mathrm{Pr}_{HQ}^{a,b,c}(R_2) = b$, $\mathrm{Pr}_{HQ}^{a,b,c}(B_1) = c$, $\mathrm{Pr}_{HQ}^{a,b,c}(B_2) = 1 - a - b - c$. Thus the set of all possible distributions HQ might have, given our assumptions, is $\mathcal{P}_{HQ} = \left\{ \mathrm{Pr}_{HQ}^{a,b,c} \mid a,b,c \geq 0,\; a+b+c \leq 1 \right\}$. In the following, we view $\mathrm{Pr}_{HQ}(R_1|R)$, $\mathrm{Pr}_{HQ}(R_1)$, $\mathrm{Pr}_{HQ}(B)$, and so on, as random variables on the space $\mathcal{P}_{HQ}$. Thus, for example, "$\mathrm{Pr}_{HQ}(R_1|R) < q$" denotes the event $\{\mathrm{Pr}_{HQ}^{a,b,c} \mid \mathrm{Pr}_{HQ}^{a,b,c}(R_1|R) < q\}$.

Again, we stress that we are not forced to use $\mathcal{P}_{HQ}$. Judy might actually have a richer model of HQ's beliefs (e.g., she might think that HQ makes finer geographical distinctions than simply the four quadrants) or a coarser model (e.g., Judy might take as the space of possibilities the possible values of $\mathrm{Pr}_{HQ}(R_1|R)$, and not reason about the rest of HQ's distribution). However, given the description of the story, $\mathcal{P}_{HQ}$ seems to be the most natural space for Judy to model her beliefs about HQ's beliefs.

Since Judy does not know what HQ actually believes, *her* beliefs will be a distribution over distributions, i.e., a distribution over $\mathcal{P}_{HQ}$. Which distribution? Again, we have many choices, but a natural one is to suppose that before Judy hears the message, she considers a uniform distribution over $(a, b, c)$ tuples. Formally, we consider the distribution function defined by $\mathrm{Pr}_{J/HQ}^{prior}\{\mathrm{Pr}_{HQ}^{a,b,c} \mid a \leq A, b \leq B, c \leq C\}) = ABC$, so that the density function is just 1. We also use the notation $\mathrm{Pr}_{J/HQ}^{q}$ to denote Judy's beliefs about HQ's beliefs after receiving $\mathbf{M}(q)$.

It might be thought that, having decided to take $\mathcal{P}_{HQ}$ as the set of possible beliefs that HQ could have and given the (implicit) assumption that Judy is initially completely ignorant of HQ's beliefs, the prior density on $\mathcal{P}_{HQ}$ is determined completely. Unfortunately, this is not the case. There is no unique "uniform distribution" on $\mathcal{P}_{HQ}$. Uniformity depends on how we choose to parameterize the space. We have chosen to parameterize the elements of this space by a triple $(a, b, c)$ denoting the probabilities of $R_1$, $R_2$, and $B_1$, respectively. However, we could have chosen to characterize an element of the space by a triple $(a', b', c')$ denoting the square of the probabilities of $R_1$, $R_2$, and $B_1$, respectively. Or perhaps more reasonably, we could have chosen to characterize an element by a triple $(a'', b'', c'')$ denoting the probability of $R$, the probability of $R_1$ given $R$, and the probability of $B_1$ given $B$. A uniform distribution with respect to either of these parameterizations would be far from uniform with respect to the parameterization we have chosen, and vice versa.[4] This is, of course, just an instance of the well-known impossibility of defining a unique notion of *uniform* in a continuous space [Howson and Urbach 1989].

Since $\mathbf{M}(q)$ is an event in the new space $\mathcal{P}_{HQ}$, Judy should be able to condition on it. One might object that, since $\mathbf{M}(q)$ is an event of measure 0, conditioning is not well defined. This is true, but there are two (closely related) ways of dealing with this problem. The more elementary and intuitive approach is based on the observation that, in practice, HQ will not (in general) be able to announce its value for $\mathrm{Pr}_{HQ}(R_1|R)$ *exactly*, since this could require HQ to announce an infinite-precision real number. It seems more reasonable to regard the announced value of $q$ as being rounded or approximated in some fashion. In particular, we might suppose that $\mathbf{M}(q)$ really means $\mathrm{Pr}_{HQ}(R_1|R) \in [q - \epsilon, q + \epsilon]$ for some small value $\epsilon > 0$. This event has non-zero probability according to $\mathrm{Pr}_{J/HQ}^{prior}$, and so conditioning is unproblematic.

---

[3]To see the possible relevance of this, note that if there are other possible messages, then the very fact that HQ's *first* message was not one of these others could be important information in and of itself: Judy might reason that $\mathbf{M}(3/4)$ must have been the most important fact HQ possessed. On the other hand, since the radio died before the message was completed, such inferences depend heavily on the protocol Judy expects HQ to follow.

[4]We note, however, that the uniform densities with respect to all 4 possible parameterizations that involving choosing 3 out of the 4 probabilities from $\mathrm{Pr}_{HQ}(R_1)$, $\mathrm{Pr}_{HQ}(R_2)$, $\mathrm{Pr}_{HQ}(B_1)$, $\mathrm{Pr}_{HQ}(B_2)$ *do* lead to the same distribution over $\mathcal{P}_{HQ}$, and so our decision to use the first three of these probabilities does not affect our analysis.

The second approach directly invokes the standard definition of conditioning on (the value of) a random variable. We briefly review the details here. Suppose we have two random variables $X$ and $Y$. If $\Pr(X = a) > 0$, then $\Pr(Y = b|X = a)$ is just defined as $\Pr(Y = b \cap X = a)/\Pr(X = a)$ as we would expect. If $\Pr(X = a) = 0$, then we take the straightforward analogue of this definition using density functions. If $f_{XY}(x,y)$ is the joint density function for $X$ and $Y$, and $f_X(x)$ is the density for $X$ alone, then the conditional density of $Y$ given $X$ is given by $f_{Y|X}(y|x) = f_{XY}(x,y)/f_X(x)$. Using the density function we can then compute the probability by integrating as usual. Further details can be found in any standard text on probability (for instance [Papoulis 1984]).

To use this approach, we need to identify a random variable $X$ and value $b$ such that $\mathbf{M}(q)$ corresponds to the event $X = b$. The choice of random variable is crucial; we can easily have two random variables $X$ and $X'$ such that $X = b$ and $X' = b$ is the same event, yet conditioning on $X = b$ and $X' = b$ leads to different results, since $X$ and $X'$ have different density functions. In our case there is an obvious choice of random variable, given our description of the situation: $\Pr_{HQ}(R_1|R)$. With this choice of random variable, it is easy to see that the two approaches give us the same answer; the use of the density function corresponds to considering a small interval around $\Pr_{HQ}(R_1|R) = q$, and then considering the limit as the interval width tends to 0.

Before computing the result of conditioning on $\mathbf{M}(q)$ (under either approach), it turns out to be useful to do some more general computations. Since $\Pr_{HQ}^{a,b,c}(R_1|R) = a/(a+b)$, we have

$$\Pr_{J/HQ}^{prior}(\Pr_{HQ}(R_1|R) < q)$$
$$= \frac{\int_{a=0}^{q} \int_{b=\frac{a(1-q)}{q}}^{1-a} \int_{c=0}^{1-a-b} 1 \, dc \, db \, da}{\int_{a=0}^{1} \int_{b=0}^{1-a} \int_{c=0}^{1-a-b} 1 \, dc \, db \, da}$$
$$= 6 \int_{a=0}^{q} \int_{b=\frac{a(1-q)}{q}}^{1-a} \int_{c=0}^{1-a-b} 1 \, dc \, db \, da$$
$$= 6 \int_{a=0}^{q} \int_{b=\frac{a(1-q)}{q}}^{1-a} (1-a-b) \, db \, da$$
$$= 6 \int_{a=0}^{q} \frac{(q-a)^2}{2q^2} \, da$$
$$= q$$

Two other results, which are derived in a similar fashion, also turn out to be useful:

$$\Pr_{J/HQ}^{prior}(\Pr_{HQ}(B) < p) = (3-2p)p^2, \text{ and} \qquad (1)$$

$$\Pr_{J/HQ}^{prior}(\Pr_{HQ}(R_1|R) < q \wedge \Pr_{HQ}(B) < p) = q(3-2p)p^2.$$

The point here is not just the values themselves, but, more importantly, that the final distribution function is the product of the first two. That is, the events

$\Pr_{HQ}(B) < p$ and $\Pr_{HQ}(R_1|R) < q$ are independent! This is of course extremely intuitive: It seems reasonable that HQ's beliefs about the probability of Judy being in Blue Territory should be independent of HQ's beliefs of her being in Red HQ area, given that she is in Red territory.

Using this, it is trivial to prove the following proposition, which holds whether we choose to use any particular $\epsilon > 0$, or if we use the standard definition of conditioning on the value of a random variable (which, as we have said, essentially corresponds to considering the limit as $\epsilon \to 0$). In this proposition, $\mathrm{pr}_B(p)$ denotes the density function for the random variable $\Pr_{HQ}(B)$; i.e., $\mathrm{pr}_B(p) = d\Pr_{J/HQ}(\Pr_{HQ}(B) < p)/dp$. Similarly, $\mathrm{pr}_B(p \mid \mathbf{M}(q)) = d\Pr_{J/HQ}(\Pr_{HQ}(B) < p \mid \mathbf{M}(q))/dp$. (Note that from (1), it immediately follows that $\mathrm{pr}_B(p) = 6p - 6p^2$, although we do not need this fact for the next result.)

**Proposition 2.1:** *In* $(\mathcal{P}_{HQ}, \Pr_{J/HQ}^{prior})$, *the events* $\Pr_{HQ}(B) = p$ *and* $\mathbf{M}(q)$ *are independent. Formally,*

$$\mathrm{pr}_B(p) = \mathrm{pr}_B(p|\mathbf{M}(q)).$$

With this result, we are very close to showing that Judy's beliefs regarding the probability of her being in Blue territory don't change as a result of the message. Notice that $\Pr_{J/HQ}^{prior}$ is a distribution on $\mathcal{P}_{HQ}$, that is, on Judy's beliefs about HQ's beliefs regarding where Judy is. The posterior distribution $\Pr_{J/HQ}^q(\cdot) = \Pr_{J/HQ}^{prior}(\cdot \mid \mathbf{M}(q))$ is still a distribution on $\mathcal{P}_{HQ}$. As a result of conditioning, Judy revises her beliefs about HQ's beliefs. But to determine Judy's beliefs about where she is we need a distribution on $\{R_1, R_2, B_1, B_2\}$. The question is how Judy's beliefs about HQ's beliefs about where she is are related to her beliefs about where she is. Notice that there is no necessary relation. After all, for all we know, Judy might think that HQ has no reliable information, and thus decide to ignore HQ's statement when forming her opinion regarding where she is. But, in keeping with the spirit of the story, we assume that Judy trusts HQ, and thus forms her beliefs by taking expectations over her beliefs about HQ's beliefs. For example, if Judy was certain that HQ was certain that she was in Blue territory, then she would ascribe probability 1 to being in Blue territory. More generally, Judy weights HQ's beliefs according to her beliefs about how likely it is that HQ holds those beliefs. We formalize this trust assumption as follows:

**(Trust)** At any point in time, Judy's beliefs about any event in the space $\{R_1, R_2, B_1, B_2\}$ are formed by taking *expectations* of HQ's probability of the same event, according to Judy's distribution over HQ's possible beliefs. Formally, we have:
$$\Pr_J^t(E)$$
$$= \int_{\Pr_{HQ}^{a,b,c} \in \mathcal{P}_{HQ}} \Pr_{J/HQ}^t(\Pr_{HQ}^{a,b,c}) \Pr_{HQ}^{a,b,c}(E) \, dabc$$

$$\left(= \int_{e=0}^{1} \mathrm{pr}_E^t(e)\, e\, de\right)$$

for $t \in \{prior\} \cup [0,1]$.(In the second line, which follows from standard probability theory, $\mathrm{pr}_E^t(e)$ is the density function of the random variable $\mathrm{Pr}_{HQ}^{a,b,c}(E)$; i.e., $\mathrm{pr}_E^t(e) = d\mathrm{Pr}_{J/HQ}^t(\mathrm{Pr}_{HQ}^{a,b,c}(E) < e)/de$.)

Note that when we apply this rule before Judy receives the message, so that $t = prior$, we have $\mathrm{Pr}_J^{prior}(B_1) = \mathrm{Pr}_J^{prior}(B_2) = \mathrm{Pr}_J^{prior}(R_1) = \mathrm{Pr}_J^{prior}(R_2) = 1/4$, which is consistent with our earlier assumption that Judy started with a uniform prior on $\{B_1, B_2, R_1, R_2\}$.

The desired result now follows quite readily using the *trust* principle after Judy has received $\mathbf{M}(q)$. The result is that, no matter what the value of $q$ is, her beliefs regarding being in Blue territory remain unchanged, exactly in accord with most people's strong intuitions.

**Theorem 2.2:** $\mathrm{Pr}_J^q(B) = 1/2$ *for all* $q \in [0,1]$.

**Proof:**
$$\mathrm{Pr}_J^q(B) = \int_{\mathrm{Pr}_{HQ}^{a,b,c} \in \mathcal{P}_{HQ}} \mathrm{Pr}_{J/HQ}^q(\mathrm{Pr}_{HQ}^{a,b,c})\, \mathrm{Pr}_{HQ}^{a,b,c}(B)\, dabc$$
$$= \int_{p=0}^{1} \mathrm{pr}_B(p \mid \mathbf{M}(q))\, p\, dp$$
$$= \int_{p=0}^{1} \mathrm{pr}_B(p)\, p\, dp \quad \text{(by Theorem 2.2)}$$
$$= \int_{p=0}^{1} (6p - 6p^2)p\, dp \quad \text{(from (1))}$$
$$= 1/2. \quad \blacksquare$$

Note that this theorem applies even if $q = 1$. Van Fraassen would interpret the message $\mathbf{M}(1)$ as meaning that Judy is definitely not in $R_2$. We interpret this it as $\mathrm{Pr}_{HQ}(R_1 | R) \in [1 - \epsilon, 1]$ for some (arbitrarily) small and unspecified $\epsilon > 0$. Although the two interpretations seem close (after all, they differ by at most $\epsilon$ in the probability that they assign to $R_1$ and $R_2$), they are *not* equivalent. As Theorem 2.2 shows, this is a significant difference. It is the claimed equivalence of the two interpretations that was behind van Fraassen's first principle, and hence his rejection of the intuitive answer that $\mathrm{Pr}_J^q(B) = 1/2$. This equivalence may be correct under van Fraassen's constraint based interpretation of $\mathbf{M}$, but it is not inevitable under our alternative reading, in which $\mathbf{M}$ is indeed a factual announcement (but about HQ's beliefs, not Judy's).

## 3    DISCUSSION

It is worth reviewing the assumptions that were necessary for us to prove Theorem 2.2. We assumed:

1. HQ's belief as to where Judy is can be characterized by a distribution on the space $\{R_1, R_2, B_1, B_2\}$.

2. Judy's beliefs regarding HQ's beliefs were characterized by the uniform distribution on HQ's beliefs, when parameterized by the tuple $(\mathrm{Pr}_{HQ}(R_1), \mathrm{Pr}_{HQ}(R_2), \mathrm{Pr}_{HQ}(B_1))$.

3. The only messages that HQ could send were those of the form $\mathbf{M}(q)$, and the one that HQ did send was the one that correctly reflected HQ's beliefs.

4. $\mathbf{M}(q)$ is interpreted as meaning that $\mathrm{Pr}_{HQ}(q) \in [q - \epsilon, q + \epsilon]$, so that we can condition on this event. (However, our result holds no matter what $\epsilon$ is. Thus we can regard $\epsilon$ as an arbitrarily small and unknown nonzero quantity.)

5. Judy's distribution on $\{B_1, B_2, R_1, R_2\}$ was derived by taking the expectation of her beliefs regarding HQ's beliefs.

Although these assumptions seem to us quite reasonable, they are clearly not the only assumptions that could have been made. It is certainly worth trying to understand to what extent our results depend on these assumptions, and in particular whether they can be extended to provide more general statements of how to update by probabilistic information.

Where does this leave CE and all the other methods considered by van Fraassen and his colleagues? As we said in the introduction, such rules may be useful in certain cases, but we believe it is an important research question to understand and explain precisely when. We do not, in particular, find CE to give particularly plausible results in the JB problem. But how could this have been predicted in advance?

The JB problem shows that we need more than just an axiomatic justification for the use of CE (or any other method of update). An alternative to the use of a rule is to do what we have done for the JB problem in this paper: that is, to try to "complete the picture" in as much detail as possible, so that ultimately all we need to do is condition. In practice, this may be unnecessarily complex and shortcuts (such as CE) might exist. However, it would be useful to understand better the assumptions that are necessary for their use to correspond to conditioning. In any case we believe that some of the issues we addressed cannot be avoided: it will never be sensible to blindly apply a rule, like CE, to all information that merely "seems" probabilistic or can be reformulated as such. Rather, one must always think carefully about precisely what the information means.

## Acknowledgments

We thank Teddy Seidenfeld and Bas van Fraassen for useful comments. The second author's work was supported in part by the NSF, under grant IRI-96-25901, and the Air Force Office of Scientific Research (AFSC), under grant F94620-96-1-0323.

## References

Bacchus, F., A. J. Grove, J. Y. Halpern, and D. Koller (1994). Generating new beliefs from old. In *Proc. Tenth Annual Conference on Uncertainty Artificial Intelligence*, pp. 37–45. Available by anonymous ftp from logos.uwaterloo.ca/pub/bacchus or via WWW at http://logos.uwaterloo.ca.

Diaconis, P. and S. L. Zabell (1982). Updating subjective probability. *Journal of the American Statistical Society* 77(380), 822–830.

Howson, C. and P. Urbach (1989). *Scientific Reasoning: The Bayesian Approach.* La Salle, Illinois: Open Court.

Jaynes, E. T. (1983). *Papers on Probability, Statistics, and Statistical Physics.* Dordrecht, Netherlands: Riedel. Edited by R. Rosenkrantz.

Jeffrey, R. C. (1983). *The Logic of Decision.* Chicago: University of Chicago Press. First Edition Published 1965.

Kullback, S. and R. A. Leibler (1951). On information and sufficiency. *Annals of Mathematical Statistics 22*, 76–86.

Papoulis, A. (1984). *Probability, Random Variables, and Stochastic Processes.* Chicago: McGraw-Hill.

Paris, J. B. and A. Vencovska (1992). A method for updating justifying minimum cross entropy. *International Journal of Approximate Reasoning 7*, 1–18.

Seidenfeld, T. (1987). Entropy and uncertainty. In I. B. MacNeill and G. J. Umphrey (Eds.), *Foundations of Statistical Inference*, pp. 259–287. Reidel. An earlier version appeared in *Philosophy of Science*, vol. 53, pp. 467–491.

Seidenfeld, T. (1997). Personal communication.

Shore, J. E. and R. W. Johnson (1980). Axiomatic derivation of the principle of maximum entropy and the principle of minimimum cross-entropy. *IEEE Transactions on Information Theory IT-26*(1), 26–37.

Uffink, J. (1995). Can the maximum entropy principle be explained as a consistency requirement? *Stud. Hist. Phil. Mod. Phys. 26*(3), 223–261.

Uffink, J. (1996). The constraint rule of the maximum entropy principle. *Stud. Hist. Phil. Mod. Phys. 27*(1), 47–79.

van Fraassen, B. C. (1980). Rational belief and probability kinematics. *Philosophy of Science 47*, 165–187.

van Fraassen, B. C. (1981). A problem for relative information minimizers. *British Journal for the Philosophy of Science 32*, 375–379.

van Fraassen, B. C. (1987). Symmetries of personal probability kinematics. In N. Rescher (Ed.), *Scientific Enquiry in Philsophical Perspective*, pp. 183–223. Lanham, Maryland: University Press of America.

van Fraassen, B. C., R. I. G. Hughes, and G. Harman (1986). A problem for relative information minimizers in probability kinematics, continued. *British Journal for the Philosophy of Science 37*, 453–475.